\documentclass[runningheads]{llncs}
\usepackage[T1]{fontenc}
\usepackage{graphics}    
\usepackage{times}       
\usepackage{amsmath}     
\usepackage{amssymb}     
\usepackage{graphicx}
\usepackage{float}
\usepackage{caption}
\usepackage{algorithm}
\usepackage[noend]{algpseudocode}
\usepackage{multirow}
\usepackage{subcaption}
\usepackage{textcomp}
\usepackage{siunitx}
\usepackage{subfig}
\usepackage{makecell}
\usepackage{url}

\usepackage[font=small]{caption}
\newcommand{\etal}{\textit{et al.}} 
\begin{document}
%
\title{Evaluating Robot Influence on Pedestrian Behavior Models for Crowd Simulation and Benchmarking}
%
\titlerunning{Social Robot Force Model}
%
\author{Subham Agrawal \inst{1} \and
Nils Dengler \inst{1, 2}\and
Maren Bennewitz \inst{1, 2, 3}
}
\institute{Humanoid Robots Lab, University of Bonn, Germany \and
The Lamarr Institute, Bonn, Germany \and
Center for Robotics, Bonn, Germany}
\authorrunning{S. Agrawal \etal}
%
\maketitle              
\begin{abstract}
The presence of robots amongst pedestrians affects them causing deviation to their trajectories. Existing methods suffer from the limitation of not being able to objectively measure this deviation in unseen cases. In order to solve this issue, we introduce a simulation framework that repetitively measures and benchmarks the deviation in trajectory of pedestrians due to robots driven by different navigation algorithms. 
We simulate the deviation behavior of the pedestrians using an enhanced Social Force Model (SFM) with a robot force component that accounts for the influence of robots on pedestrian behavior, resulting in the Social Robot Force Model (SRFM). Parameters for this model are learned using the pedestrian trajectories from the JRDB dataset \cite{martin2021jrdb}.
Pedestrians are then simulated using the SRFM with and without the robot force component to objectively measure the deviation to their trajectory caused by the robot in 5 different scenarios.
Our work in this paper is a proof of concept that shows objectively measuring the pedestrian reaction to robot is possible. We use our simulation to train two different RL policies and evaluate them against traditional navigation models.

\end{abstract}
\section{Introduction}
\vspace{-5pt}
The increasing presence of robots in everyday life has made the issue of social navigation of mobile robots among pedestrians more relevant than ever. 
As robots become more integrated into various public and private spaces, from shopping malls \cite{niemela2017monitoring,niemela2019social}, to hospitals \cite{kyrarini2021survey} and homes \cite{gates2007robot,henschel2021makes}, their ability to navigate in crowded environments efficiently and safely, but also adhere to social norms, becomes crucial.

\begin{figure}[t!]
	\centering
	\includegraphics[width=0.75\linewidth]{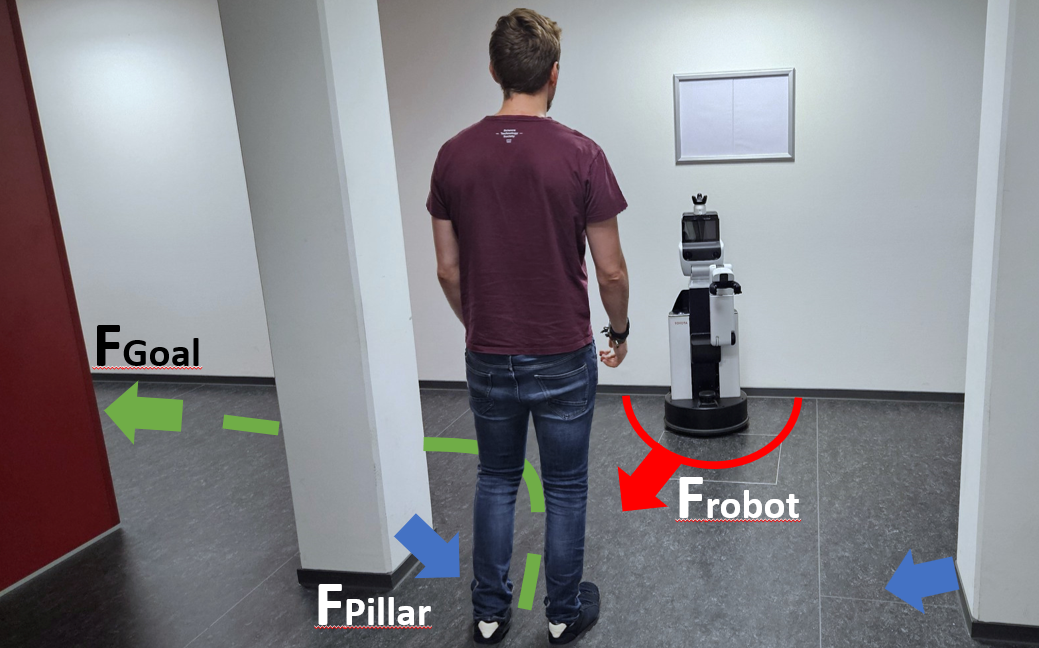}
	\caption{Example scenario of a human navigating to a goal. While many of the forces used in the social force model are well studied, such as attraction to the goal (green arrow) or repulsion from obstacles, such as the pillar (blue arrow), the robot force (red arrow) is not well understood, even though it significantly affects human behavior.
	} 
	\label{fig:cover}
\end{figure}

Research in the area of social robot navigation attempts to address this challenge and focuses on two main parts: the social aspect and the robot navigation. 
While robot navigation has been extensively studied and solved using both traditional and data-driven approaches, a repetitive measure for the affect of robots on pedestrians remains a dynamic area of research \cite{mavrogiannis2023core,hirose2023sacson}. 

As more methods are developed to navigate robots socially amongst humans, the need to evaluate these methods becomes paramount. 
Over the years, the evaluation of social navigation methods has evolved to include more nuanced metrics that account for the mutual influence between robots and pedestrians. 
Early evaluations primarily focused on the accuracy of pedestrian trajectory predictions and the efficiency of robot navigation using metrics such as arrival rate, path length, collision rate, average time to goal, etc.
However, contemporary evaluations increasingly consider the social acceptability of robot behavior and its adaptability in dynamic environments. 
Primarily subjective evaluation methods are used for gauging this while there exists still no consensus on which measures to use as standards \cite{mavrogiannis2023core}. 

Recently, Hirose~\etal~\cite{hirose2023sacson} suggested a metric for measuring effect of robots on pedestrians by evaluating the counterfactual perturbation of human trajectories caused by the presence of a robot in the scene.
However, the dynamic application of this metric remains a challenge, especially when relying on pre-recorded datasets, which limit the ability to test policies in scenarios different from those in the dataset. 

\begin{figure}[t!]
	\centering
	\includegraphics[width=0.9\linewidth]{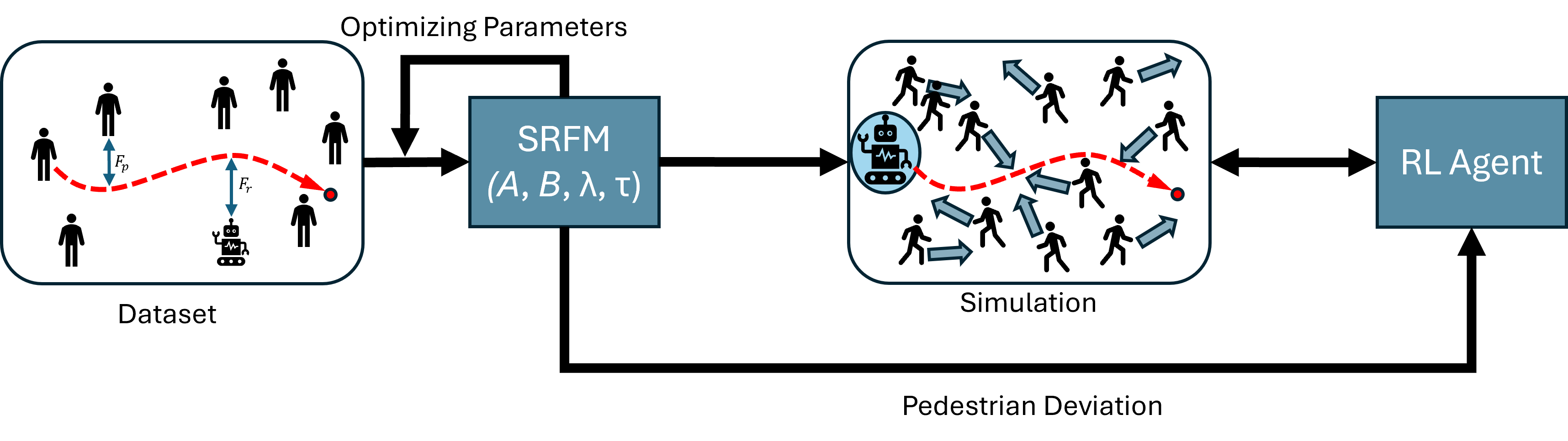}
	\caption{Architecture of the simulation system. The Social Robot Force Model (SRFM) parameters are learned from the dataset. The learned SRFM is then used to drive the pedestrians in simulation while an RL agent drives the robot during training and evaluation phases.} 
	\label{fig:architecture}
\end{figure}

To address these issues, we present a simulation framework designed to repetitively and objectively measure the effect of robots driven by different social navigation algorithms on pedestrian trajectory for evaluation and benchmarking.
In order to achieve this, we use an additional force-factor in the social force model~(SFM)~\mbox{\cite{helbing1995social,regier2019improving}}, inspired by the work done by Ferrer \etal~\cite{ferrer2013robot}, to model the robot's influence on the pedestrian walking behavior without modeling it as a simple obstacle as shown in Fig \ref{fig:cover}. 
This social robot force model (SRFM) is used to account for changes in the pedestrians' behavior that cannot be modeled adequately by the original forces of the SFM.
Our framework, as shown in Fig \ref{fig:architecture}, simulates pedestrian behavior based on real-world datasets according to our SRFM, and allows for the creation of scenarios not covered in the original dataset. This flexibility enables the development of navigation policies for specific scenarios such as complex hallways as well as generalized scenarios such as pedestrian crossings.
To validate this claim, we developed and evaluated a reinforcement learning (RL) policy that uses SRFM to learn a human-aware robot navigation behavior by minimizing the influence of robot force on the pedestrian trajectory. The results show that the learned policy causes least deviation to pedestrian trajectory amongst compared model in most of the evaluation scenarios.

\section{Related Work}
\vspace{-5pt}
Since the first deployment of the robots RHINO \cite{burgard1998} and Minerva \cite{thrun2000probabilistic}, research in the domain of social navigation in robotics has primarily revolved around three core approaches: traditional model-based methods, data-driven learning approaches, and hybrid methods combining both elements.

Traditional model-based methods use predefined physical laws to explain and predict pedestrian motion and in turn make navigation choices for the robot. 
Well known and studied traditional methods are the Social Force Model (SFM) \cite{helbing1995social}, Velocity Obstacle (VO) approach \cite{large2002using}, continuum theory \cite{hughes2002continuum}, as well as the dynamic window approach (DWA) \cite{fox1997dynamic}. 
According to the SFM, pedestrian motion results from the sum of various forces acting on them at any time, including attraction to goals and repulsion from obstacles and other pedestrians. 
VO is a collision avoidance technique that  looks at the permissible velocities for the robot that do not fall in the velocity cone of the obstacle, which prevent it from colliding against static as well as dynamic obstacles in its environment while navigating towards its goals. 
DWA is a real-time local path planning algorithm that dynamically adjusts a robot's velocity within a specified window to avoid collisions and select the safest path \cite{fox1997dynamic,missura2019predictive}. These methods are easy to implement.
However, the parameters used with these models need to be hand-tuned and can vary. Some models such as the continuum theory \cite{hughes2002continuum} are also restricted, as they could be used to predict the general movement of a particular density of crowd but fail to accurately model the interactions on the level of individual pedestrians.

Recent machine learning methods involve the collection of extensive datasets capturing pedestrian and robot interactions, which are then used to train predictive models. 
The models learn to forecast pedestrian behavior, guide robots in real-time, and depending on the comparison scenario, can outperform traditional methods \cite{mavrogiannis2023core}.


Hirose~\etal~\cite{hirose2023sacson} argued the measure of social ability of a robot's navigation algorithm being the ability to cause the least perturbation in their original trajectory. We rely on this idea and incorporate it in a simulation framework which allows objective and repetitive measurement of this deviation in a pedestrian's trajectory caused by the robot.


\section{Our Approach}


In this section, we introduce the extended social force model that includes the repulsion effect of navigating robots on pedestrians. We explain in detail our methods for learning the parameters of the social force, defining an objective metric to repetitively measure trajectory deviation of pedestrians in the presence of a navigating robot, setting up a simulation environment for training a navigation policy, and evaluating a policy against other state-of-the-art algorithms using this simulation environment. 
\subsection{Social Robot Force}
\label{sec:SRFM}
The original SFM \cite{helbing1995social} uses three major forces - attraction towards the goal $f_a$, repulsion from other pedestrians in the vicinity $f_p$, and repulsion from obstacles in the surroundings $f_o$:
\ref{eq:2}. 
\begin{equation}
\label{eq:1}
F = f_a + f_p + f_o
\end{equation}

Our modified SRFM, uses an additional force component - repulsion from the robot $f_r$: 
\begin{equation}
\label{eq:2}
F = f_a + f_p + f_o + f_r
\end{equation}

The different force components are calculated as follows. 
The attraction force towards the goal $f_a$ is described in Eq. \ref{eq:3}:
\begin{equation}
\label{eq:3}
f_a = \frac{v_i - v_0}{\tau}
\end{equation}
$\tau$, representing the time a pedestrian takes to adjust their current velocity $v_{i}$ to match their desired velocity towards the goal $v_{0}$, has to be learned.
The repulsion forces from pedestrians $f_p$, obstacles $f_o$, and robots $f_r$ share the same formula as depicted in Eq. \ref{eq:4}: 
\begin{equation}
\label{eq:4}
f_p = Ae^\frac{d-x}{B}\psi
\end{equation}
In this function, the parameters to be learned are $A$, indicating the strength of the force, and $B$, representing the distance from which the force starts to have a significant effect.
Additionally, an anisotropic value $\psi$ is used to show that pedestrians experience stronger repulsive forces from those in front of them (using the angle between the pedestrians - $\phi$):
\begin{equation}
\label{eq:5}
\psi = \lambda + (1 - \lambda)\frac{(1 + cos(\phi))}{2}
\end{equation}
This force decreases to the sides and becomes zero behind the pedestrian. 
This factor $\psi$ is included as an extra multiplicative factor in the force from Eq. \ref{eq:4} and helps in more accurately modeling pedestrian behavior. The value of $\lambda$ is another parameter that needs to be learned from real-world data.

\subsection{Learning SFRM parameters}

\begin{table}[ht!]
    \centering
    \caption{Parameter values for the Social Robot Force Model compared to \cite{ferrer2013robot}}
    \begin{tabular}{|c|cccccc|} \hline 
        Paper & $A_p$ & $B_p$ & $\lambda$ & $\tau$ & $A_r$ & $B_r$ \\ \hline 
        Ours  & 2 & 0.89 & 0.4 & 0.6 & 7.93 & 0.99 \\  
        Ferrer \etal \cite{ferrer2013robot} & 2.66 & 0.79 & 0.59 & 0.43 & 2.66 & 0.79 \\ \hline
    \end{tabular}
    \label{tab:SFRM_parameters}
\end{table}

In this work, we learn all parameters based on trajectories of the JRDB dataset \cite{martin2021jrdb}. 
Since we wanted to use annotated values of trajectories, we filtered only those where there were no obstacles, as the dataset does not have annotated information of the obstacles.
In the experiments presented, the modified SFM does not take into account the force from obstacles due to this limitation of the dataset.
However, this factor is well studied and can be added to the equation for further research.
Therefore, the modified model represents pedestrian motion in open spaces such as public crossings and wide footpaths which involve pedestrian-pedestrian interactions and have a negligible chance of pedestrian-obstacle interactions.
To learn the factors of the SFM independent from the robot force, we processed the dataset to categorize its trajectories into interaction and non-interaction types.
Non-interaction category involve the trajectories of pedestrians far away from the robot (greater than 3 meters) and visually not affected by it. 
These trajectories are used to learn the parameters for the pedestrian repulsion component. 
For the repulsion force from robots, only parameters $A$ and $B$ need to be learned from the interaction trajectories, which represent those trajectories where pedestrians are near the robot and its presence causes deviation in their path.
These trajectories are used to learn the parameters for the robot force component while retaining the values for pedestrian part from the non-interaction trajectories. 
We used a non-linear least squares optimization technique from the SciPy package \cite{2020SciPy-NMeth} to learn the parameters using the data for both the scenarios. The resulting parameter values are summarized in Tab. \ref{tab:SFRM_parameters} and are shown in comparison to the values found by Ferrer \etal \cite{ferrer2013robot}.

\subsection{Deviation Metric and Simulation Scenarios}
\label{sec:sim}

To effectively benchmark social navigation policies, realistic and repeatable scenarios are needed as well as a well defined objective metric.
Therefore, for training and evaluation of navigation policies we utilize the learned Social Robot Force Model to create different scenarios as described in the following.

\vspace{-10pt}

\subsubsection{Training}
As training environment, we define a free space of $15\times15$ meters within our simulation.
The robot's start and goal positions are randomly generated while maintaining a minimum distance of 5 meters. 
During training, 10 pedestrians are randomly sampled within the environment boundaries while maintaining a minimum distance of $2m$ (the robot's social zone) from the robot's start position. 
In addition, each pedestrian's goal is randomly sampled, restricting it to be outside the robot's goal's social zone and keeping a minimum distance of 7 meters from that pedestrian's initial location. 
In order to provide a dynamic and crowded training environment, once a pedestrian has reached its goal, but the training episode has not ended, it is assigned a new goal as described above.
\vspace{-10pt}
\subsubsection{Evaluation}
To evaluate our learned policy after training, we use five different scenarios, the first three based on scenarios described by Francis~\etal~\cite{francis2023principles} while the remaining ones as variations based on extreme situations. 

\begin{figure}[t]
    \centering 
\begin{subfigure}{0.3\textwidth}
  \includegraphics[width=\linewidth]{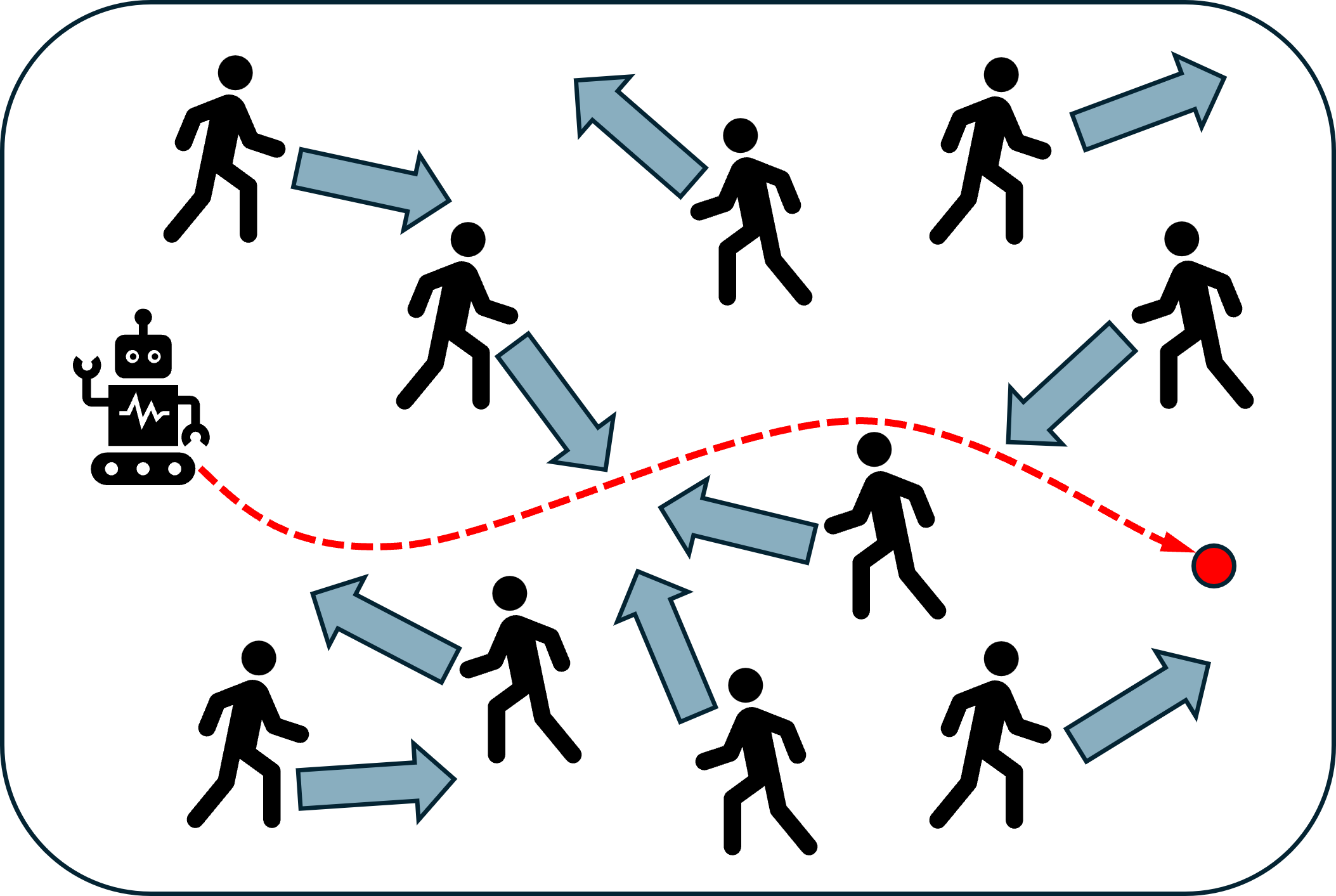}
  \caption{Scenario: Random}
  \label{fig:scenario_random}
\end{subfigure}\hfil 
\begin{subfigure}{0.3\textwidth}
  \includegraphics[width=\linewidth]{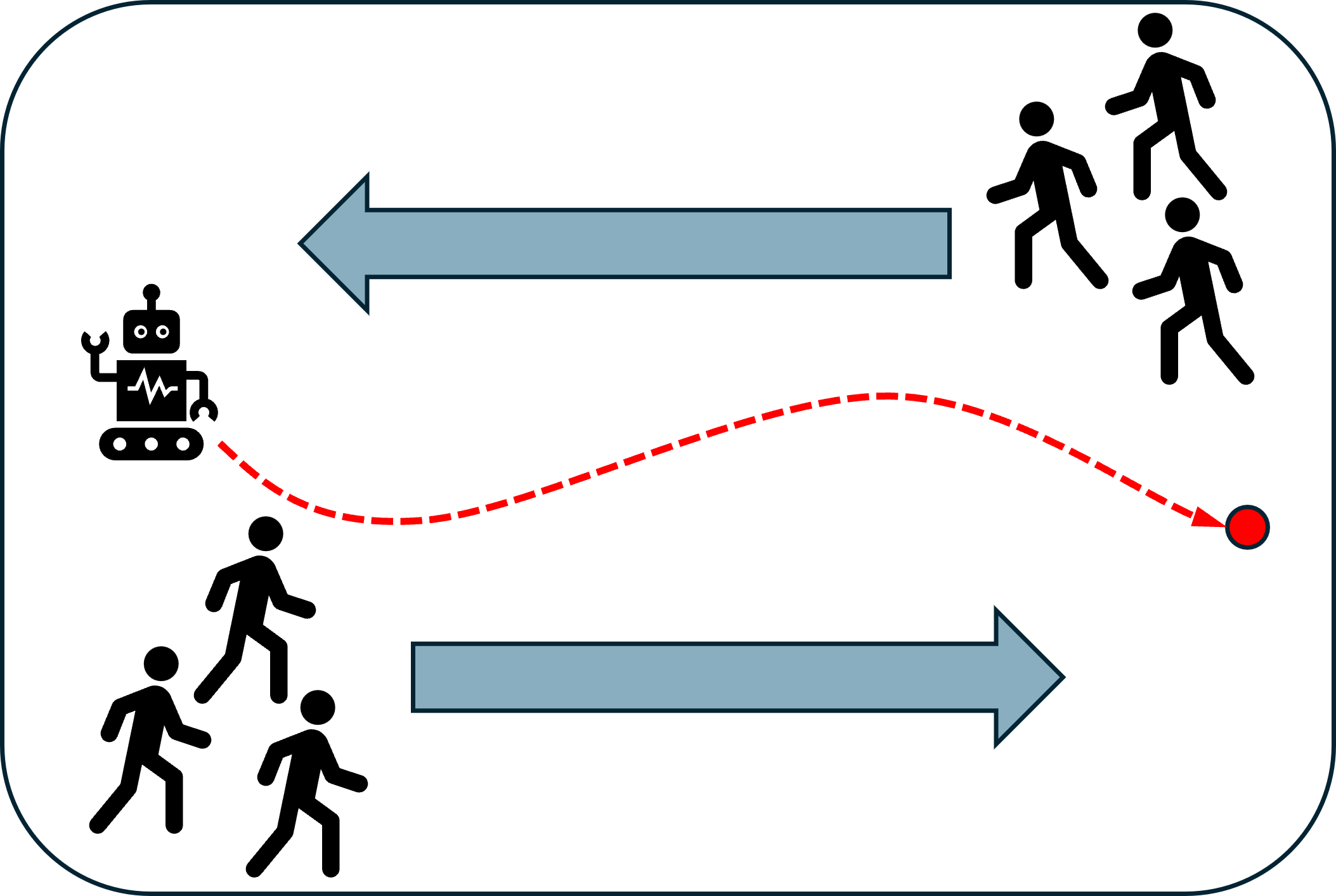}
  \caption{Scenario1: Footpath}
  \label{fig:scenario_1}
\end{subfigure}\hfil 
\begin{subfigure}{0.3\textwidth}
  \includegraphics[width=\linewidth]{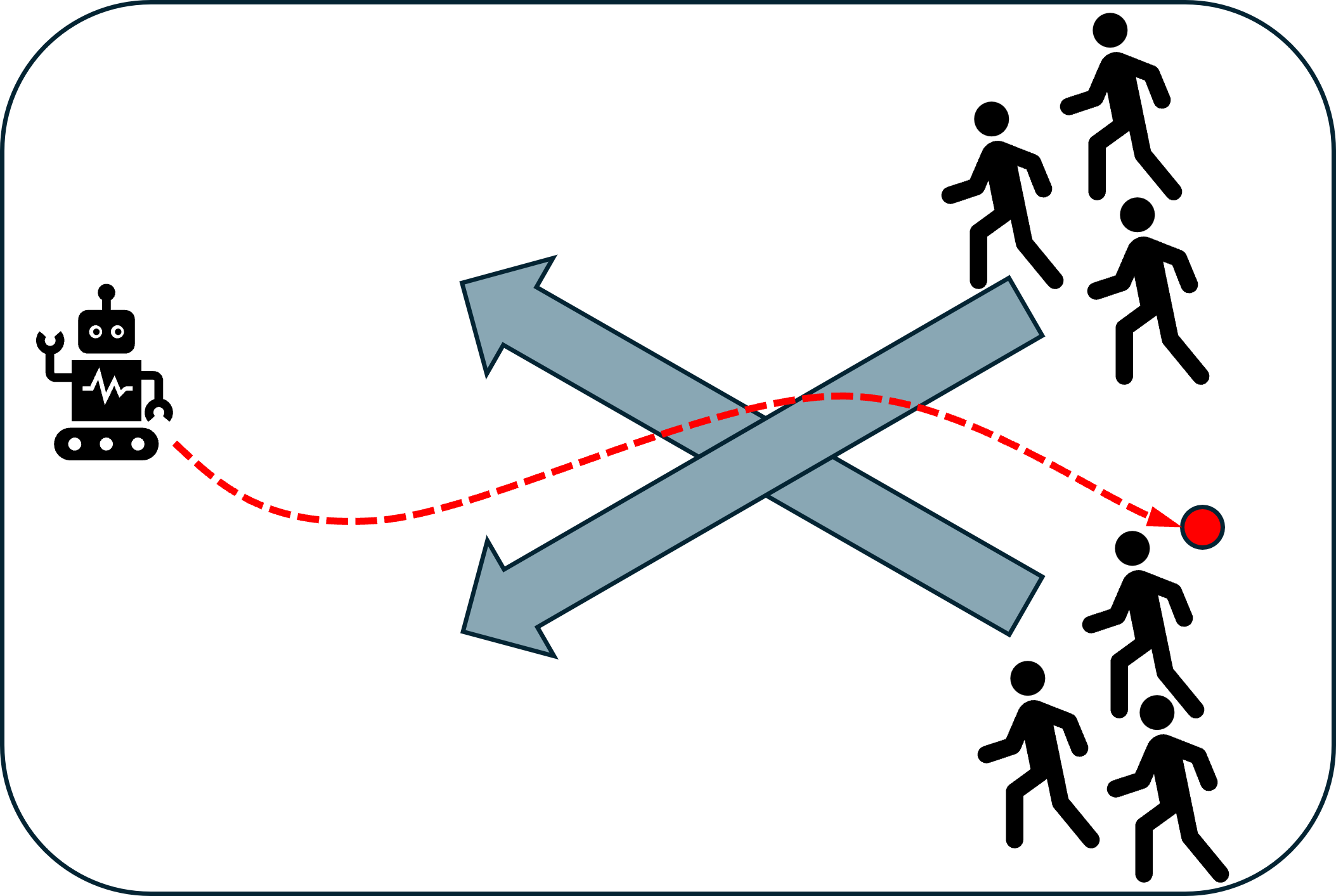}
  \caption{Scenario2: Crosswalk}
  \label{fig:scenario_2}
\end{subfigure}

\medskip
\begin{subfigure}{0.3\textwidth}
  \includegraphics[width=\linewidth]{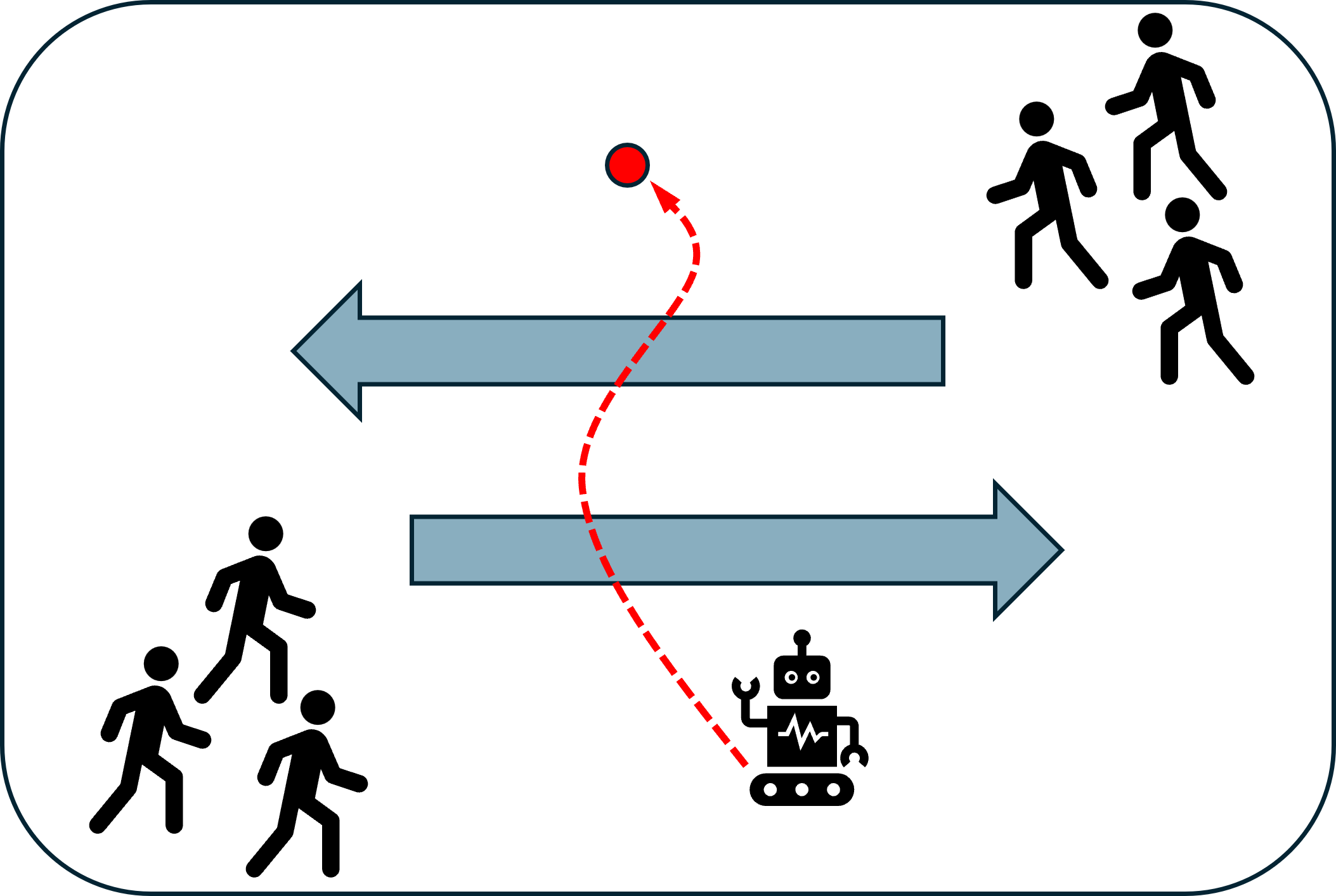}
  \caption{Scenario3: Crossfootpath}
  \label{fig:scenario_3}
\end{subfigure}\hfil 
\begin{subfigure}{0.3\textwidth}
  \includegraphics[width=\linewidth]{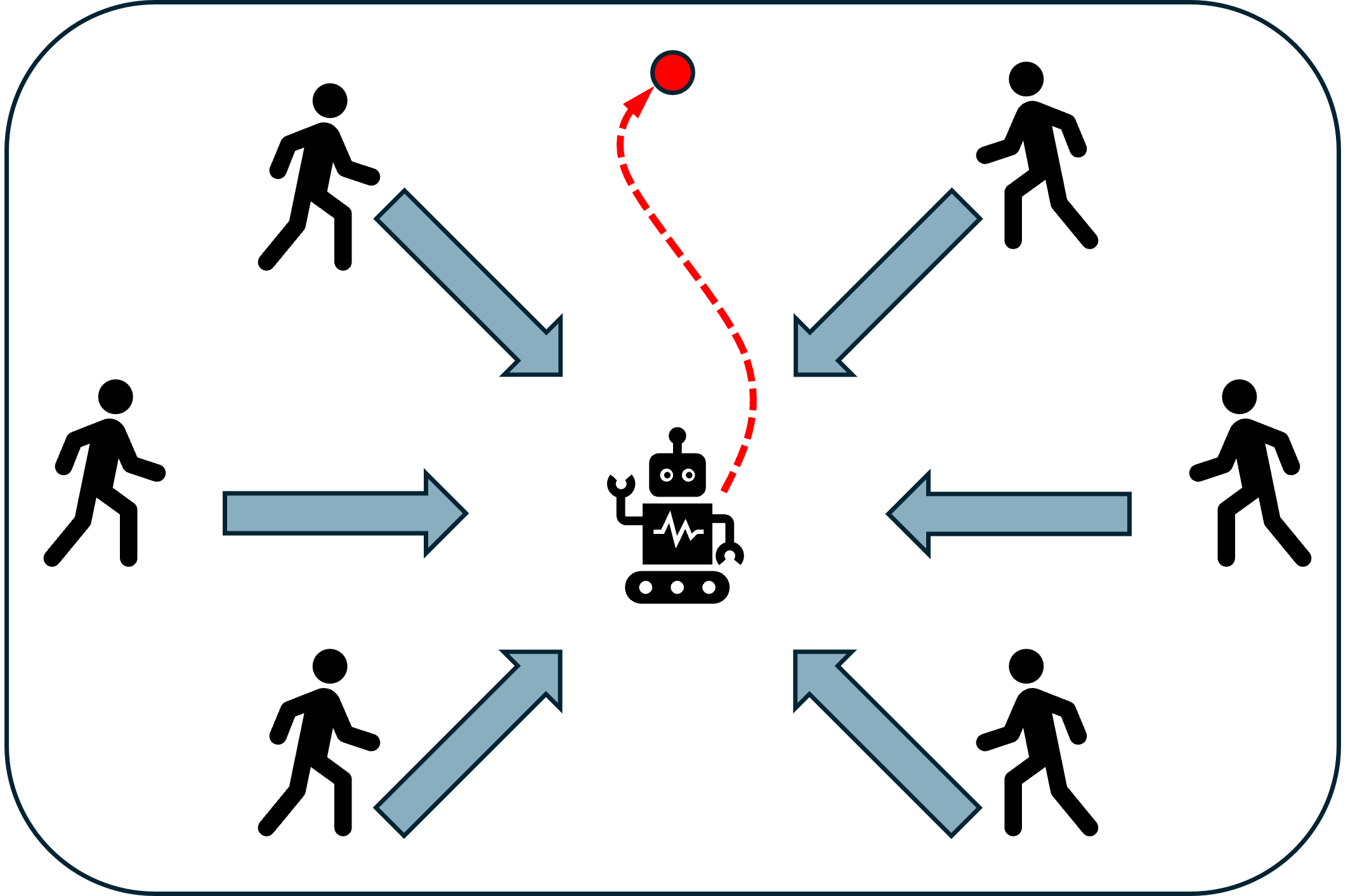}
  \caption{Scenario4: Box}
  \label{fig:scenario_4}
\end{subfigure}\hfil 
\begin{subfigure}{0.3\textwidth}
  \includegraphics[width=\linewidth]{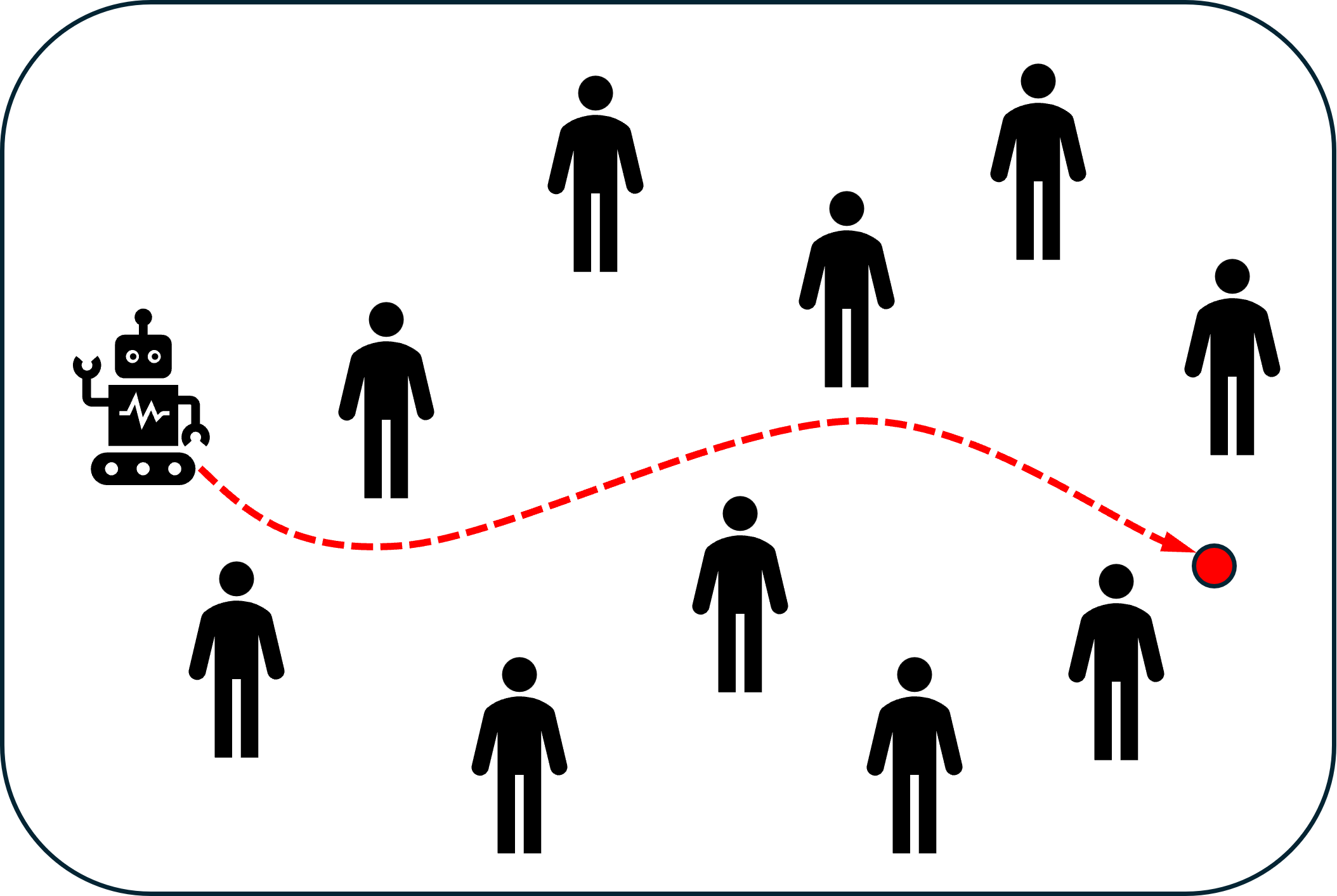}
  \caption{Scenario5: Concert}
  \label{fig:scenario_5}
\end{subfigure}
\caption{Overview of the different scenarios used for the evaluation of navigation policies. (a) Training environment that is used for the RL agent. (b), (c), and (d) visualize the three common pedestrian streams as described in \cite{francis2023principles}. (e) and (f) visualize two extreme scenarios.}
\label{fig:scenarios}
\end{figure}

In \textbf{Scenario 1}, we simulate a sidewalk or similar walking area by maintaining the parallel flow of pedestrians bypassing each other. For this scenario, the start and end positions of the pedestrians are set to create a natural parallel flow, as can be seen in Fig. \ref{fig:scenario_1}. The start and end positions of the robot are fixed and shown as a red circle and a green star.

For \textbf{Scenario 2} a pedestrian crossing situation is created. 
The pedestrian start and goal positions are set to form a natural cross-flow, as shown in \ref{fig:scenario_2}. The robot start and goal positions are set up to directly disrupt the flow at the pedestrian crossing point, making it more difficult for the robot to minimize its influence on the pedestrian trajectories.

\textbf{Scenario 3} is a variation of Scenario 1 where the robot has to cut across the parallel flow of pedestrians. This scenario tests the navigation policy on its timing as the correct timing can minimize pedestrian deviation.

In \textbf{Scenario 4}, the robot starts in the middle of a formation where it is surrounded be pedestrians standing in a circle and trying to reach the point diametrically opposite to them. This creates an aggressive scenario where the robot has to face incoming pedestrians from all directions.

\textbf{Scenario 5} aims to simulate a minimal motion scenario where pedestrians have the same start and end positions. This is representative of a concert or other public events where pedestrians do not change their positions. The position of the start and end goal of the robot in this scenario presents a situation where it has a high chance of cutting across the crowd inadvertently causing deviation in their positions.

\subsubsection{Trajectory deviation metric}

A major feature of our benchmark simulation is the objective metric to measure the deviation in trajectory of the pedestrians caused by the presence of robots driven by social navigation algorithms. 
This metric is based on the prior research done by Hirose~\etal~\cite{hirose2023sacson}.
The SRFM consists of various individual force components, making it efficient to evaluate a social navigation policy twice under the same scenario, with one key difference: the robot's impact on the pedestrian. 
To measure the counterfactual perturbations caused by the robot on the pedestrian's trajectory, we first set up the SRFM with its robot force component active and run the benchmark to record pedestrian trajectories.
We then conduct a second run of the deterministic benchmark using the SRFM but with the robot force component disabled. This provides us with pedestrian trajectories for the same scenario but without the robot's influence.
By comparing the differences between these two sets of trajectories and quantifying this difference using Fréchet distance \cite{deheuvel2023iros} which is a metric that reports the similarity between two curves, we can evaluate the deviation caused by the robot driven by a social navigation policy. 
The main idea of our benchmark is that the path which a pedestrian takes due to presence of the robot is different from how the person would have walked if the robot was not there.
However, one thing to note is that pedestrian behavior is not limited to pure repulsive behavior. Pedestrians also tend to show actively engaging or neutral behavior towards the robot which will be considered in future versions of the benchmark system.

\subsubsection{Other metrics}
Apart from the above, three other popular metrics are evaluated and recorded as part of the current simulation system. These are:
\begin{itemize}
    \item \textbf{Minimum Robot Distance:} The minimum distance between the robot and any pedestrian during that particular evaluation scenario.
    \item \textbf{Trajectory Length:} The average of total distance travelled by the pedestrians during the scenario.
    \item \textbf{Trajectory Time:} The average of total time taken by the pedestrians to traverse to their goal during the scenario.
\end{itemize}

\subsection{Learning a Navigation Policy}
\label{sec:rl}
To demonstrate the advantage of the SRFM and our simulation framework, we use reinforcement learning (RL) to train a policy that navigates the robot through a crowd of pedestrians.
While reaching a certain goal in the environment, the robot is tasked to cause the least disturbance to the pedestrians velocity and trajectory.

RL can be framed as a control problem modeled by a partially observable Markov decision process (POMDP). 
In this framework, the agent cannot directly observe its exact state \( s_t \) at time step \( t \); instead, it relies on the current observation \( o_t \) to approximate the state, such that \( s_t \approx o_t(s_{t-1}, a_t) \) where \( s_{t-1} \) is the previous state and \( a_t \) is the current action. 
The objective is to discover a stochastic policy \( \pi(a_t | o_t) \) that maximizes the expected reward \( R \) over an episode, with \( T \) representing the number of time steps and \( \gamma \) being a discount factor. 
The optimization goal is expressed as:
\begin{center}
\begin{equation}
\label{eq:6}
    \max \mathbb{E} \left( \sum_{t=0}^T \gamma^t R(s_t, a_t) \right)
\end{equation}
\end{center}

In the following we elaborate more on the action, observation and reward functions that we used to train and evaluate the RL agent in the context of SRFM and social norms.

\subsubsection{Actions}
We use a continuous action space for the agent that consists of linear and angular velocities $(v, w)$, with a range of $v \in [-0.5, 0.5]$ \SI{}{\metre\per\second} and $w~\in~[-\pi, \pi]$~\SI{}{\radian\per\second}. 
\subsubsection{Observations}
All components are either relative to the robot's position or a boolean.
As policy observation of the environment, we use the distance and angle to the navigation goal as well as for each pedestrian with in a predefined social zone, as described in Sec. \ref{sec:SRFM}.
We use an upper bound of the ten closest pedestrians, as other research suggests an upper bound of interfering humans within a scene of nine \cite{hossain2022sfmgnet}. For scenarios with less than ten pedestrians within the robot's social zone, the observation is padded with vectors of zeros.  
We also include the last action taken by the robot $(v, w)$, as well as the success and termination criteria (boolean) in the observation function.
\subsubsection{Reward}
We keep the reward as simple as possible to encourage faster training, since it is an important part for training convergence.
Our reward function $r_{total}$ consists of three components,
\begin{equation}
\label{eq:7}
    r_{total} = r_{term} + - k_1 \cdot r_{dist} - k_2 \cdot r_{div},
\end{equation}
with $k_1, k_2$ as scaling factors.
$r_{term}$ is a large, sparse termination reward, which is positive when the episode is successful, and otherwise negative. Additionally, we penalize collisions with obstacles double as much as reaching the maximum time step limit.
$r_{dist}$ is the Euclidean distance to the goal position, normalized to the range $[0,1]$ by dividing it with the distance when the environment is reset at the start of every episode.
To learn from our SRFM implementation, we use $r_{div}$ to penalize the agent for causing any divergence to the pedestrian from its original path. This value is obtained by calculating the next step of the pedestrian using SRFM with and without robot force and getting the Euclidean distance between these two predicted positions - original position predicted without robot force and deviated position predicted with robot force.

\subsubsection{Training method}
To train the policy, we use the Twin-Delayed Deep Deterministic Policy Gradient (TD3) algorithm implementation of stable baselines 3 \cite{stable-baselines3} with a maximum of $2e6$ training steps and a learning rate of $1e-4$. As TD3 is an off-policy RL algorithm, we set the length of its experience replay buffer to $1e6$. 
Each episode consists of 750 steps, where start and goal positions of the robot and the pedestrians are assigned randomly while considering a certain minimum distance $d=2$ between any of them to avoid unlikely or dangerous situations where the robot spawns on top of or very close to a pedestrian or vice versa. 

\section{Experiments and Results}
In this section, we first evaluate the influence of the SRFM in terms of prediction accuracy on a given dataset. Second, we demonstrate its usability for evaluating robot navigation policies.

\subsection{SRFM Prediction Accuracy}


The learned parameters for the SRFM were tested on a subset of trajectories derived from the JRDB dataset. 
We are going to refer to the force components of the SFRM as attraction force, pedestrian force, and robot force.
The testing was done in three phases.
First, we tested the accuracy of our learned pedestrian force on pedestrian trajectories that are not disrupted by the robot (non-interaction), which resulted in an Average Displacement Error ($ADE$) of $0.70m$. This metric is used to measure the average of the absolute distance in the predicted and actual position of the agent over the course of the entire trajectory.
Next, we checked the accuracy of our learned pedestrian force on pedestrian trajectories that are disrupted by the robot (interaction) and consider the robot as a pedestrian as well. The ADE in this case increased to $0.75m$ showing that pedestrians try to maintain a different distance from the robot than other pedestrians.
Finally, for the interaction trajectories, we considered the robot as a separate entity and introduced the robot force. The ADE with this change improved to $0.59m$. This is because pedestrian trajectories in interaction scenarios deviate away from the robot and using robot force leads to the driven pedestrian following a similar trajectory. The results are presented in Tab. \ref{tab:ADE_table}.

\begin{table}[t!]
    \centering
    \caption{Error in trajectory prediction on the JRDB dataset between the social force model and our social robot force model. As can be seen from the results, using the social force model parameters for the robot increases the average displacement error whereas the learned robot force component from social robot force model significantly reduces it.}
    \begin{tabular}{|c|ccc|}
    \hline
        Model & Force & Test Trajectory & ADE\\ \hline
        SFM & $f_a + f_p (pedestrians)$ & Non-interaction & 0.70m \\
        SFM & $f_a + f_p (pedestrians) + f_p (robot)$ & Interaction & 0.75m \\
        SRFM & $f_a + f_p (pedestrians) + f_r (robot)$ & Interaction & \textbf{0.59m} \\
    \hline
    \end{tabular}
    \label{tab:ADE_table}
\end{table}

\begin{table*}[ht!]
\setlength\tabcolsep{0pt}
\centering
\caption{Performance of our policy against a $RL_{base}$, DWA, and VO in terms of the average Fréchet distance, trajectory length, the time taken to complete the trajectory and the minimum distance from a pedestrian to the robot. All values are averaged over 100 runs with $\ast$ indicating significance compared to ours according to the independent t-test with p=0.05. Results show our policy outperforms others in Scenarios 1,2,3, and 5 with significant results in Scenarios 1 and 5.}
\begin{tabular}{|c|c|c|c|c|c|}
\Xhline{4\arrayrulewidth}
  &\hspace{.5em}\textbf{Approach} \hspace{.5em} & \hspace{.5em}\textbf{Fréchet Dist. \textdownarrow} \hspace{.5em}& \hspace{.5em}\textbf{Min. Robot Dist. \textuparrow} \hspace{.5em}&  \hspace{.5em}\textbf{Traj. Length \textdownarrow} \hspace{.5em}& \hspace{.5em}\textbf{Time \textdownarrow} \hspace{.5em}\\
\Xhline{4\arrayrulewidth}
\multirow{4}{*}{\textbf{Scenario 1} }    &    Ours    &    $\ast \textbf{0.69} \pm \textbf{0.05}$    &    $1.42 \pm 0.09$    &    $8.94 \pm 0.10$    &    $18.71 \pm 0.15$    \\
\cline{2-6}
    &    $RL_{base}$    &    $0.78 \pm 0.04$    &    $1.49 \pm 0.18$    &    $9.00 \pm 0.13$    &    $18.71 \pm 0.20$    \\
 \cline{2-6}
    &    DWA    &    $0.76 \pm 0.05$    &    $1.55 \pm 0.15$    &    $\ast \textbf{8.77} \pm \textbf{0.14}$    &    $\ast \textbf{14.49} \pm \textbf{1.20}$    \\
 \cline{2-6}
    &    VO    &    $0.76 \pm 0.04$    &    $\ast \textbf{1.63} \pm \textbf{0.20}$    &    $9.02 \pm 0.10$    &     $23.01 \pm 1.27$    \\
\Xhline{4\arrayrulewidth}
\multirow{4}{*}{\textbf{Scenario 2} } & Ours              & $\textbf{1.25} \pm \textbf{0.17}$ & $1.42 \pm 0.15$                        & $14.63 \pm 0.12$                       & $19.57 \pm 0.70$ \\
\cline{2-6}
 & $RL_{base}$          & $1.32 \pm 0.14$                        & $1.40 \pm 0.19$                        & $14.74 \pm 0.23$                       & $20.74 \pm 0.54$                        \\
 \cline{2-6}
 & DWA               & $1.57 \pm 0.17$                        & $1.58 \pm 0.05$ & $\ast \textbf{14.09} \pm \textbf{0.38}$                       & $\ast \textbf{17.60} \pm \textbf{1.71}$                        \\
 \cline{2-6}
 & VO                & $1.29 \pm 0.15$                        & $\textbf{1.61} \pm \textbf{0.05}$                        & $14.63 \pm 0.17$     & $21.12 \pm 2.53$                        \\
\Xhline{4\arrayrulewidth}
\multirow{4}{*}{\textbf{Scenario 3} } & Ours              & $\textbf{0.68} \pm \textbf{0.08}$      & $\ast \textbf{1.76} \pm \textbf{0.08}$ & $8.68 \pm 0.08$             & $19.11 \pm 0.44$      \\
\cline{2-6}
 & $RL_{base}$          & $0.72 \pm 0.09$                        & $1.62 \pm 0.17$                        & $8.79 \pm 0.11$                        & $19.37 \pm 0.41$                        \\
 \cline{2-6}
 & DWA               & $0.75 \pm 0.08$                        & $1.70 \pm 0.13$                        & $\textbf{8.67} \pm \textbf{0.14}$                        & $\ast \textbf{14.82} \pm \textbf{1.28}$                        \\
 \cline{2-6}
 & VO                & $0.76 \pm 0.11$                        & $1.69 \pm 0.07$                        & $8.86 \pm 0.15$                        & $20.01 \pm 0.81$                        \\
\Xhline{4\arrayrulewidth}
\multirow{4}{*}{\textbf{Scenario 4} } & Ours              & $4.72 \pm 0.36$                        & $1.52 \pm 0.04$                        & $8.68 \pm 0.35$ & $13.00 \pm 0.02$ \\
\cline{2-6}
 & $RL_{base}$          & $4.43 \pm 0.45$                        & $1.53 \pm 0.12$                        & $9.06 \pm 0.42$                        & $14.19 \pm 0.72$                        \\
 \cline{2-6}
 & DWA               & $5.67 \pm 0.65$ & $1.69 \pm 0.05$                        & $\ast \textbf{7.38} \pm \textbf{0.91}$                        & $\ast \textbf{11.65} \pm \textbf{1.48}$                        \\
 \cline{2-6}
 & VO                & $\ast \textbf{4.38} \pm \textbf{0.38}$                        & $\ast \textbf{1.72} \pm \textbf{0.04}$ & $9.09 \pm 0.38$                        & $14.60 \pm 0.57$                        \\
\Xhline{4\arrayrulewidth}
\multirow{4}{*}{\textbf{Scenario 5} } & Ours              & $\ast \textbf{0.70} \pm \textbf{0.05}$      & $1.56 \pm 0.10$                        & $\ast \textbf{2.18} \pm \textbf{0.13}$                        & $\ast \textbf{22.02} \pm \textbf{0.10}$ \\
\cline{2-6}
 & $RL_{base}$          & $0.79 \pm 0.10$                        & $1.57 \pm 0.10$                        & $2.34 \pm 0.20$ & $22.23 \pm 0.21$                        \\
 \cline{2-6}
 & DWA               & $0.81 \pm 0.06$                        & $\ast \textbf{1.66} \pm \textbf{0.06}$ & $2.64 \pm 0.27$                        & $33.91 \pm 7.21$                        \\
 \cline{2-6}
 & VO                & $0.79 \pm 0.07$                        & $1.62 \pm 0.08$                        & $2.39 \pm 0.17$                        & $25.15 \pm 2.74$                        \\ 
\Xhline{4\arrayrulewidth}
\end{tabular}
\label{tab:results}
\end{table*}

\subsection{Baselines}
To see the influence of the SRFM on our RL-agent, we trained another agent in the same way as described in Sec. \ref{sec:rl}, neglecting the path deviation penalty ($r_{div}$) for pedestrians within the social zone of 3m.
While this agent still avoids pedestrians, it does not incorporate any knowledge of its influence on their trajectories. In the following we refer to this baseline as $RL_{base}$.
Furthermore, we implement DWA \cite{fox1997dynamic} and VO \cite{large2002using} as baselines of traditional methods to compare against our agent. 
The pedestrian detection range of the robot is kept at 2m and the parameters are tuned to achieve similar average velocity as the trained agent. 
However, VO was not able to finish Scenarios 3, 4, and 5 with these parameters.
Hence, the detection range is reduced to 1.4m which allows VO to calculate feasible velocities for all scenarios. 
\subsection{Results}
The results of the evaluation in all presented scenarios are shown in Tab. \ref{tab:results}.
As can be seen, our trained agent outperforms the baseline algorithms w.r.t Fréchet distance in Scenarios 1, 2, 3, and 5. 
This behaviour is expected since these scenarios represent general and solvable situations of pedestrian dynamics, where there are paths that the robot can take while causing the least amount of disruption to pedestrians. 
Since our trained agent is specifically tuned to optimize its navigation around this behavior, the results show the best performance compared to the other algorithms in the Fréchet distance metric.

Due to its higher complexity in comparison to the others, Scenario 4 has to be discussed in more detail.
In Scenario 4, which requires the robot to navigate from in between a constantly closing circle of pedestrians, the VO algorithm performs the best, likely because it is able to find the optimum trajectory to avoid the incoming pedestrians which our agent was not trained to do. However, it is interesting to note that the performance of VO is not significantly better than the $RL_{base}$ model which was trained to also not care about pedestrian deviation.

These results show that our simulation system provides an adequate environment to train and evaluate agents that can perform on par or even outperform traditional navigation algorithms in our proposed benchmark metric and can also be evaluated on other established benchmark metrics.


\subsection{Future Directions}
In this work we propose the Fréchet distance as objective measure for deviation in pedestrians trajectory.
Please note that this is not the only or best metric to evaluate social compliance, as deviation in pedestrian trajectory can also be caused when the pedestrian is interested in the robot and comes close to it rather than avoiding it. 
The pedestrian may also be neutral to the presence of the robot and thus not show any deviation in trajectory. 
These cases however are observed in lesser frequency in the dataset used compared to the deviation caused due to avoiding the robot and will be explored in future work.

\section{Conclusion}
In this paper, we introduced a simulation framework to objectively measure and benchmark the deviation in the trajectory of pedestrians due to robots driven by different navigation algorithms. 
By extending the traditional Social Force Model (SFM) to include robot influence on pedestrian behavior, our Social Robot Force Model (SRFM) offers enhanced prediction accuracy for pedestrian trajectories disrupted by robots. 
Experiments showed a low Average Displacement Error (ADE) for the prediction accuracy of the SRFM, and our reinforcement learning policy trained with SRFM demonstrated improved results causing less deviation to pedestrian trajectories. The code for the work done in this paper can be found in \url{https://github.com/HumanoidsBonn/SRFM-Pedestrian-Deviation-Benchmark}.
\\\\
\begin{credits}
\ackname: This work has been partially funded by the German research foundation (DFG) under the grant number BE 4420/2-2 (FOR 2535 Anticipating Human Behavior), the Federal Ministry of Education and Research~(BMBF) under the grant number 16KIS1949, and within the Robotics Institute Germany, grant No. 16ME0999.
\end{credits}
%
%
%
\bibliographystyle{splncs04}
\bibliography{bibliography}

\begin{thebibliography}{10}
\providecommand{\url}[1]{\texttt{#1}}
\providecommand{\urlprefix}{URL }
\providecommand{\doi}[1]{https://doi.org/#1}

\bibitem{martin2021jrdb}
Martin-Martin, R., Patel, M., Rezatofighi, H., Shenoi, A., Gwak, J., Frankel, E., Sadeghian, A., Savarese, S.: Jrdb: A dataset and benchmark of egocentric robot visual perception of humans in built environments. IEEE transactions on pattern analysis and machine intelligence  (2021)

\bibitem{niemela2017monitoring}
Niemel{\"a}, M., Arvola, A., Aaltonen, I.: Monitoring the acceptance of a social service robot in a shopping mall: First results. In: Proceedings of the Companion of the 2017 ACM/IEEE International Conference on Human-robot Interaction (2017)

\bibitem{niemela2019social}
Niemel{\"a}, M., Heikkil{\"a}, P., Lammi, H., Oksman, V.: A social robot in a shopping mall: studies on acceptance and stakeholder expectations. Social robots: Technological, societal and ethical aspects of human-robot interaction  (2019)

\bibitem{kyrarini2021survey}
Kyrarini, M., Lygerakis, F., Rajavenkatanarayanan, A., Sevastopoulos, C., Nambiappan, H.R., Chaitanya, K.K., Babu, A.R., Mathew, J., Makedon, F.: A survey of robots in healthcare. Technologies  (2021)

\bibitem{gates2007robot}
Gates, B.: A robot in every home. Scientific American  (2007)

\bibitem{henschel2021makes}
Henschel, A., Laban, G., Cross, E.S.: What makes a robot social? a review of social robots from science fiction to a home or hospital near you. Current Robotics Reports  (2021)

\bibitem{mavrogiannis2023core}
Mavrogiannis, C., Baldini, F., Wang, A., Zhao, D., Trautman, P., Steinfeld, A., Oh, J.: Core challenges of social robot navigation: A survey. ACM Transactions on Human-Robot Interaction  (2023)

\bibitem{hirose2023sacson}
Hirose, N., Shah, D., Sridhar, A., Levine, S.: Sacson: Scalable autonomous control for social navigation. IEEE Robotics and Automation Letters  (2023)

\bibitem{helbing1995social}
Helbing, D., Molnar, P.: Social force model for pedestrian dynamics. Physical review E  (1995)

\bibitem{regier2019improving}
Regier, P., Shareef, I., Bennewitz, M.: Improving navigation with the social force model by learning a neural network controller in pedestrian crowds. In: 2019 European Conference on Mobile Robots (ECMR). IEEE (2019)

\bibitem{ferrer2013robot}
Ferrer, G., Garrell, A., Sanfeliu, A.: Robot companion: A social-force based approach with human awareness-navigation in crowded environments. In: 2013 IEEE/RSJ International Conference on Intelligent Robots and Systems. IEEE (2013)

\bibitem{burgard1998}
Burgard, W., Cremers, A., Fox, D., Hähnel, D., Lakemeyer, G., Schulz, D., Steiner, W., Thrun, S.: The interactive museum tour-guide robot (1998)

\bibitem{thrun2000probabilistic}
Thrun, S., Beetz, M., Bennewitz, M., Burgard, W., Cremers, A.B., Dellaert, F., Fox, D., Haehnel, D., Rosenberg, C., Roy, N., et~al.: Probabilistic algorithms and the interactive museum tour-guide robot minerva. The international journal of robotics research  (2000)

\bibitem{large2002using}
Large, F., Sckhavat, S., Shiller, Z., Laugier, C.: Using non-linear velocity obstacles to plan motions in a dynamic environment. In: 7th International Conference on Control, Automation, Robotics and Vision, 2002. ICARCV 2002. IEEE (2002)

\bibitem{hughes2002continuum}
Hughes, R.L.: A continuum theory for the flow of pedestrians. Transportation Research Part B: Methodological  (2002)

\bibitem{fox1997dynamic}
Fox, D., Burgard, W., Thrun, S.: The dynamic window approach to collision avoidance. IEEE Robotics \& Automation Magazine  (1997)

\bibitem{missura2019predictive}
Missura, M., Bennewitz, M.: Predictive collision avoidance for the dynamic window approach. In: 2019 International Conference on Robotics and Automation (ICRA). IEEE (2019)

\bibitem{2020SciPy-NMeth}
Virtanen, P., Gommers, R., Oliphant, T.E., Haberland, M., Reddy, T., Cournapeau, D., Burovski, E., Peterson, P., Weckesser, W., Bright, J., {van der Walt}, S.J., Brett, M., Wilson, J., Millman, K.J., Mayorov, N., Nelson, A.R.J., Jones, E., Kern, R., Larson, E., Carey, C.J., Polat, {\.I}., Feng, Y., Moore, E.W., {VanderPlas}, J., Laxalde, D., Perktold, J., Cimrman, R., Henriksen, I., Quintero, E.A., Harris, C.R., Archibald, A.M., Ribeiro, A.H., Pedregosa, F., {van Mulbregt}, P., {SciPy 1.0 Contributors}: {{SciPy} 1.0: Fundamental Algorithms for Scientific Computing in Python}. Nature Methods  (2020)

\bibitem{francis2023principles}
Francis, A., P{\'e}rez-d'Arpino, C., Li, C., Xia, F., Alahi, A., Alami, R., Bera, A., Biswas, A., Biswas, J., Chandra, R., et~al.: Principles and guidelines for evaluating social robot navigation algorithms. arXiv preprint arXiv:2306.16740  (2023)

\bibitem{deheuvel2023iros}
de~Heuvel, J., Corral, N., Kreis, B., Conradi, J., Driemel, A., Bennewitz, M.: Learning depth vision-based personalized robot navigation from dynamic demonstrations in virtual reality  (2023)

\bibitem{hossain2022sfmgnet}
Hossain, S., Johora, F.T., M{\"u}ller, J.P., Hartmann, S., Reinhardt, A.: Sfmgnet: A physics-based neural network to predict pedestrian trajectories. arXiv preprint arXiv:2202.02791  (2022)

\bibitem{stable-baselines3}
Raffin, A., Hill, A., Gleave, A., Kanervisto, A., Ernestus, M., Dormann, N.: Stable-baselines3: Reliable reinforcement learning implementations. Journal of Machine Learning Research  (2021)

\end{thebibliography}
%




\end{document}